\documentclass[letterpaper]{article} 
\usepackage{aaai24}  
\usepackage{times}  
\usepackage{helvet}  
\usepackage{courier}  
\usepackage[hyphens]{url}  
\usepackage{graphicx} 
\usepackage{natbib}  
\usepackage{caption} 
\usepackage{algorithm}
\usepackage{algorithmic}
\usepackage{booktabs}
\usepackage{multirow}
\usepackage{indentfirst}
\usepackage{tabularx}
\usepackage{array}
\usepackage{amsmath}
\usepackage{graphicx} 
\usepackage{subcaption} 
\usepackage{newfloat}
\usepackage{listings}
\DeclareCaptionStyle{ruled}{labelfont=normalfont,labelsep=colon,strut=off} 
\floatstyle{ruled}
\newfloat{listing}{tb}{lst}{}
\floatname{listing}{Listing}
\title{FMPAF: How Do Fed Chairs Affect the Financial Market?\\ A Fine-grained Monetary Policy Analysis Framework on Their Language}
\author {
    Yayue Deng\textsuperscript{\rm 1},
    Mohan Xu\textsuperscript{\rm 2},
    Yao Tang\textsuperscript{\rm 2,}\thanks{Yao Tang is the corresponding author.}
}
\affiliations {
    \textsuperscript{\rm 1}Beijing University of Posts and Telecommunications, Beijing, China\\
    \textsuperscript{\rm 2}Peking University, Beijing, China\\
    yayue.deng@bupt.edu.cn, xu.mohan@stu.pku.edu.cn, ytang@gsm.pku.edu.cn
}

\usepackage{bibentry}

\begin{document}

\maketitle

\begin{abstract}
The effectiveness of central bank communication is a crucial aspect of monetary policy transmission. While recent research has examined the influence of policy communication by the chairs of the Federal Reserve on various financial variables, much of the literature relies on rule-based or dictionary-based methods in parsing the language of the chairs, leaving nuanced information about policy stance contained in nonverbal emotion out of the analysis. In the current study, we propose the Fine-Grained Monetary Policy Analysis Framework (FMPAF), a novel approach that integrates large language models (LLMs) with regression analysis to provide a comprehensive analysis of the impact of the press-conference communications of chairs of the Federal Reserve on financial markets. We conduct extensive comparisons of model performance under different levels of granularity, modalities, and communication scenarios. Based on our preferred specification, a one-unit increase in the sentiment score is associated with an increase of the price of S\&P 500 Exchange-Traded Fund by approximately 500 basis points, a 15-basis-point decrease in the policy interest rate, while not leading to a significant response in exchange rates.

\end{abstract}
\section{Introduction}
Volatility in the financial market is driven by numerous factors, such as macroeconomic policies, public confidence, and news releases. The complex relationship among these variables renders the forecast of financial variables exceedingly challenging, as effective forecasting depends heavily on the gathering and parsing of a vast array of external information. Among various sources, information released by central banks is particularly significant, as their decisions and guidance have a profound and widespread impact on the economy and the financial market \cite{tsukioka2020tone, bennani2020does, trilliondollar}. For instance, in August 2020, Jerome H. Powell, the Chairman of the Federal Reserve (the Fed), delivered a high-profile speech at the Jackson Hole Symposium, where he announced a major shift in the Fed's approach toward inflation. Powell indicated that the Fed would be willing to permit inflation to exceed the conventional 2\% target for a period of time before hiking interest rates. This announcement led to immediate and notable reactions in the stock market. The S\&P 500 and Nasdaq indices surged to a record high on impact, as investors interpreted the Fed's current stance as a signal for prolonged low-interest rates, which typically benefit the stock market.

Because central banks are well known to exert heavy influence on the financial market, several AI-driven works in the finance literature \cite{thevoice,monopoly,trilliondollar,matsui2021using,DBLP:conf/ranlp/ZirnMS15,frunza2020information} concentrate on central banks' policy communication with the public, especially communication at the Fed's Federal Open Market Committee (FOMC) meeting. A recent work \cite{monopoly} constructs a multimodal dataset, MONOPOLY, from press conference videos for the purpose of financial forecasting. \citet{matsui2021using} propose a word embedding-based framework for understanding the policy communication of the Fed. \citet{DBLP:conf/ranlp/ZirnMS15} use graph clustering methods to unsupervisedly reveal opinion groups in transcripts of the FOMC discussions. \citet{frunza2020information} introduce an unsupervised approach to extract economic concept-value pairs from FOMC statements. However, these methods either utilize traditional dictionary-based sentiment analysis \cite{tsukioka2020tone,narain2023market}, or only focus on the pre-processing or construction of data sources \cite{trilliondollar,frunza2020information,monopoly}, without analyzing the effects of information on the financial market. The proposed methods are often not validated with empirical tests.

In terms of the first limitation of sentiment analysis, the challenge in understanding policy stance lies in quantifying its subtle cues, which are often soft and implicit information. Previous studies mainly employ dictionary-based or rule-based methods, attempting to quantify textual data through pre-defined emotion lexicons or rules, respectively. Dictionary-based sentiment analysis suffers from a limited vocabulary and inaccurate performance compared to large language models (LLMs). Besides, the complexity of policy stances extends far beyond the dimensions of emotion, necessitating more nuanced analysis methods for a comprehensive understanding. Secondly, the analysis of financial market responses is under-explored in these studies \cite{DBLP:conf/ranlp/ZirnMS15,frunza2020information,monopoly,trilliondollar}, as only classification accuracy is reported. In addition, \citet{rozkrut2007quest} points out that the effects of central banks' communication on the financial market vary with the sources of communication, such as press conference post FOMC meetings and speech by Fed officials at other venues. Therefore, it is valuable to investigate the effect of different sources of communication on macroeconomic variables.

To address the limitations discussed above, we introduce a Fine-grained Monetary Policy Analysis Framework (FMPAF) that adopts a powerful pre-trained language model (PLMs) to enhance comprehension capability and fine-tune it in the monetary policy stance corpus to transfer learning ability to the specific policy domain. Leveraging the FMPAF, our research delves into the financial market responses to Fed communication from different sources, including our personally collected semi-annual monetary policy hearings and press conferences. Moreover, we employ a policy indicator to elucidate the impulse response of stock prices, exchange rates, and interest rates in the financial markets of the US and other important economies.


The current study is closely related to \citet{thevoice}, which also focuses on the policy communication of the Fed in the form of textual and acoustic materials, but their study relies on analysis at the conversation level. Such a coarse-grained analysis can lead to the loss of information. In contrast, we employ fine-grained sentiment analysis and conduct multi-modal analysis at the sentence level. Our comparative analysis reveals that our fine-grained method enhances predictability, revealing a significant correlation with financial variables.


In summary, the proposed FMPAF aims to establish a correlation between shifts in policy sentiment and subsequent market movements by quantifying sentiment in the communications made by chairs of the Fed. Our framework applies PLMs-based models from both the textual and acoustic sides to detect the fine-grained policy stance embedded in the language of the chairs. Moreover, our empirical tests suggest a high degree of validity for the FMPAF, as the framework generates a reliable indicator of the chairs' underlying policy stance. Extensive comprehensive analysis demonstrates that the indicator of the policy stance is predictive of the fluctuation of the financial market.

 
\begin{figure*}[ht!]
  \centering
  \includegraphics[width=\linewidth]{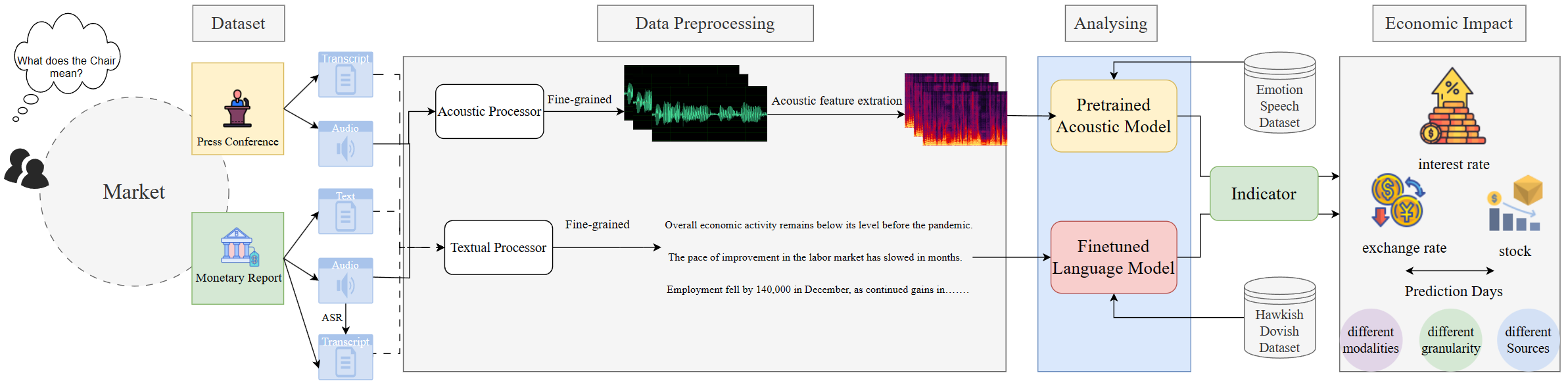}
  \caption{Overview of our proposed Policy Analysis Framework}
  \label{pic:overview}
\end{figure*}

\section{Background and Related Work}
\subsection{Sentiment Analysis in Finance}
\label{subsec:sentiment_analysis}
The term \textit{Sentiment Analysis} was first defined by Nasukawa et al. \cite{Sentimentanalysis} as determining the subjectivity polarity (positive or negative) and polarity strength (strongly positive, mildly positive, weakly positive, etc.) of a given review text. Sentiment analysis is increasingly recognized as a pivotal task within the financial domain, closely linked with market sentiment, which subsequently influences price movements. Early literature \cite{emodict2,thevoice} predominantly focused on word categorization methods, such as the bag of words technique, to assess tonality. The introduction of the polarity lexicon by \cite{Mcdonaldwords} stands out as a significant advancement in the development of sentiment analysis methodologies for finance. Further, \cite{financewords} contributes new lexicons of positive and negative terms specifically for financial contexts. Nonetheless, these dictionary-based approaches face two principal challenges: 1) Context Insensitivity: They often neglect the subtleties of context, such as sarcasm or culturally specific expressions, resulting in imprecise sentiment analysis; 2) Inadequacy in Processing Complex Emotions: These methods are limited in their ability to interpret complex emotional expressions and fail to accurately discern sentiment intensities or nuances.
Recently, large language models like Bert \cite{bert} and RoBERTa \cite{roberta} have demonstrated extensive capabilities, significantly enhancing a range of natural language processing (NLP) tasks. Despite their remarkable efficacy in a wide array of NLP applications, Pretrained Language Models (PLMs) encounter limitations in domain-specific tasks. This is primarily due to the marked disparity between the distribution of general training data and the specialized language of specific domains \cite{DBLP:conf/acl/GururanganMSLBD20,DBLP:journals/health/GuTCLULNGP22}. An expression that holds a strong semantic orientation in one domain might necessitate a different interpretation in another. To mitigate this, several financial language models \cite{finbert,finbert2} have been proposed to enhance performance in analyzing financial and economic texts. For example, recent endeavors, such as by \cite{trilliondollar}, have concentrated on monetary orientation analysis through the fine-tuning of large language models.
\subsection{Policy Communication of Central Bank}
Economists have long recognized the importance of central bank communication \cite{kohn2003central}, making efforts to study its impact on financial markets both quantitatively and qualitatively \cite{blinder2008central,blinder2018through,neuhierl2019monetary,coenen2017communication}. Despite all these efforts, the communication mechanism between the Fed Chairs and the market, an important channel of central bank communication, has not been fully explored and understood due to the technical limitations. Most related papers are restricted to the impact of the Chairs' communication on the domestic financial market \cite{thevoice, hansen2016shocking}. To fill in the gap, our paper investigates the impact of the Chairs' language on the international financial market for the very first time, which contributes to the broad literature on cross-border transmission of monetary policy.

\begin{table*}[ht!]
\centering
\caption{Summary Statistics of Collected Datasets: Policy Reports and Press Conferences}
\label{tab:data}
\renewcommand{\arraystretch}{1.15} 
\resizebox{0.96\textwidth}{!}{
\begin{tabular}{c|c|c|c|cccc}
\toprule
\textbf{Source} & \textbf{Years} & \textbf{File} & \textbf{Modalities} & \textbf{File} & \textbf{Avg. Words} & \textbf{Avg. Duration (s)} & \textbf{Total (words \& s)} \\ 
\midrule
\multicolumn{8}{c}{Panel A: Raw Data} \\
\midrule
\multirow{3}{*}{Policy Reports} & \multirow{3}{*}{2003-2023} & Readout & Textual & 42 & 2376.1 & - & 92708 \\
 & & Speeches & Acoustic & 30 & - & 9018.6 & 261540.4 \\
 & & Transcribed text & Textual & 30 & 21059.5 & - & 610726.0 \\
 \midrule
\multirow{2}{*}{Press Conferences} & \multirow{2}{*}{2011-2022} &  Speeches & Acoustic & 47 & - & 3379.3 & 160845.6 \\
 & & Transcribed text & Textual & 47 & 8445.4 & - & 405379.0 \\
 \midrule
\multicolumn{8}{c}{Panel B: Processed Data} \\
\midrule
\multirow{3}{*}{Policy Reports} & \multirow{3}{*}{2003-2023} & Readout & Textual & 3230 & 28.7 & - & 92708 \\
 & &  Speeches & Acoustic & 36203 & - & 6.1 & 219963.3 \\
 & & Transcribed text & Textual & 36210 & 16.8 & - & 610726.0 \\
 \midrule
\multirow{2}{*}{Press Conferences} & \multirow{2}{*}{2011-2022} &  Speeches & Acoustic & 19025 & - & 8.2 & 158826.5 \\
 & & Transcribed text & Textual & 19440 & 20.4 & - & 405379.0 \\
\bottomrule
\end{tabular}}
\end{table*}

\section{Methodology}
As shown in Fig~\ref{pic:overview}, the proposed FMPAF contains four parts: 1) Dataset Collection, 2) Preprocessing, 3) Deep-learning-based multi-modal Analysing, and 4) Economic impact analysis via indicators. In this section, we first describe the background of our data resource. Besides, we illustrate our deep-learning-based method to conduct acoustic and linguistic analysis for investigating the economic impact of policy communication.
\subsection{Problem formulation}
In order to investigate the impact of the Chairman's communication on financial markets, we formulate the problem as a regression task. Following \cite{jorda2005estimation}, the impulse response of financial market and the Fed's communication can be formulated as the local projection specification as follows:
\begin{equation}
\begin{split}
&\text{Outcome}_{t,t+h} = b_0^{(h)} + b_1^{(h)} \text{Sentiment}_{t}\\
&\quad + b_2^{(h)} \text{FFRShock}_{t} + b_3^{(h)} \text{FGShock}_{t} + b_4^{(h)} \text{APShock}_{t} \\
&\quad  + b_5^{(h)} \text{ShadowRate}_{t} + \text{error}_{t}^{(h)},   
\end{split}   
\label{equ:regression}
\end{equation}
where $t$ denotes the date of the press conferences, $Sentiment_t$ measures the policy stance embedded in the opening remarks and Q\&A sessions, $FFRShock_t$, $FGShock_t$, $APShock_t$ and $ShadowRate$ capture the policy action of the Fed. 
\begin{table}[h]
\centering
\caption{Summary of Control Variables and Sources}
\label{tab:controls}
\begin{tabular}{@{}cc@{}}
\toprule
    \textbf{Controls}           & \textbf{Sources}           \\ 
\midrule
    FFR Shock          & \multirow{3}{*}{Swanson (2021)}       \\
    FG Shock           &                                       \\
    AP Shock           &                                       \\
    \midrule
    Shadow Rate & Wu and Xia (2016) \\ 
\bottomrule
\end{tabular}
\end{table}
To avoid the omitted variable problem, we introduce four control variables into our regression, as shown in Table~\ref{tab:controls}. Therefore, the impact of $sentiment_t$ on financial outcomes can be clearly identified.   
For the financial outcomes, we employ stock prices, exchange rates, and interest rates, using the log difference between close and open prices to measure the accumulated return, which is referred to as the $Outcome_{t,t+h}$. Take the representative stock price in the USA (SPY) as an example, we compute the $Outcome_{t,t+h}$ according to: 
\begin{equation}
\begin{split}
 \text{Outcome}_{t,t+h}^{SPY} = \log(\text{SPY}_{t+h}^{\text{close}}) - \log(\text{SPY}_t^{\text{open}})   
\end{split}
\end{equation}

We estimate the above specification for each horizon in a daily frequency, ranging from 0 day to 15 days. To illustrate the dynamic effects explicitly, we draw the estimated coefficients. 90 \% bias-corrected confidence intervals are reported, shown by the dash lines in each figure.

\subsection{Dataset Collection and Preprocessing}
Our dataset comprises two primary types of data:
\begin{enumerate}
\item \textbf{The Fed's monetary policy hearings}: presented to Congress by the Chair on monetary policy and economic outlook, including both a written report and an oral testimony at a Congressional hearing, conducted formally and systematically.
\item \textbf{The Fed's press conferences}: held after FOMC meetings, where the Chair announces monetary policy decisions and answers journalists' questions.
\end{enumerate}
Both types of data are essential for how the Fed communicates its monetary policy to the public and legislative bodies.

\textbf{Monetary Policy Hearings}
Our dataset of this part includes samples from the United States Senate Committee on Banking, Housing, and Urban Affairs website\footnote{https://www.banking.senate.gov/hearings/}, detailing hearings on financial and economic policies. An example is the July 13, 2017\footnote{https://www.banking.senate.gov/hearings/the-semiannual-monetary-policy-report-to-the-congress}, hearing on the semiannual monetary policy report, featuring Janet L. Yellen, then Chair of the Fed, as the key witness. The website provides dates, times, locations, and additional resources like opening remarks and testimonies.

In our study, we scraped the United States Senate Committee on Banking, Housing, and Urban Affairs website to compile a list of all hearings. We then applied two criteria to identify the Monetary Policy Hearings: 1) the title must contain “monetary policy report,” and 2) the “Chair” must be listed among the attendees. From this, we extracted essential details from the hearing URLs, such as primary attendees, audio recordings, the chair's testimonies, and meeting dates. For the formal statement documents (readouts) reported to Congress, we used Pdfplumber \cite{pdfplumber}, a Python package, to extract the text and segment it into sentences. For meetings without transcripts, we employed an Automatic Speech Recognition (ASR) tool, WhisperX \cite{whisperx}, for transcription. Our final dataset, as summarized in Table~\ref{tab:data}, includes 42 Monetary Policy Meetings from 2003 to 2023, each with a PDF of the readout and, when available, an MP3 audio file. Notably, audio recordings were not available for all meetings; 12 out of 42 lacked audio files. We segmented a total of 36,203 audio pieces, averaging about 6.1 seconds each, mostly ranging from 3 to 6 seconds, showing a uniform distribution in duration.

\begin{figure*}
  \centering
  \begin{subfigure}[b]{0.5\linewidth}
    \centering
    \includegraphics[width=\linewidth]{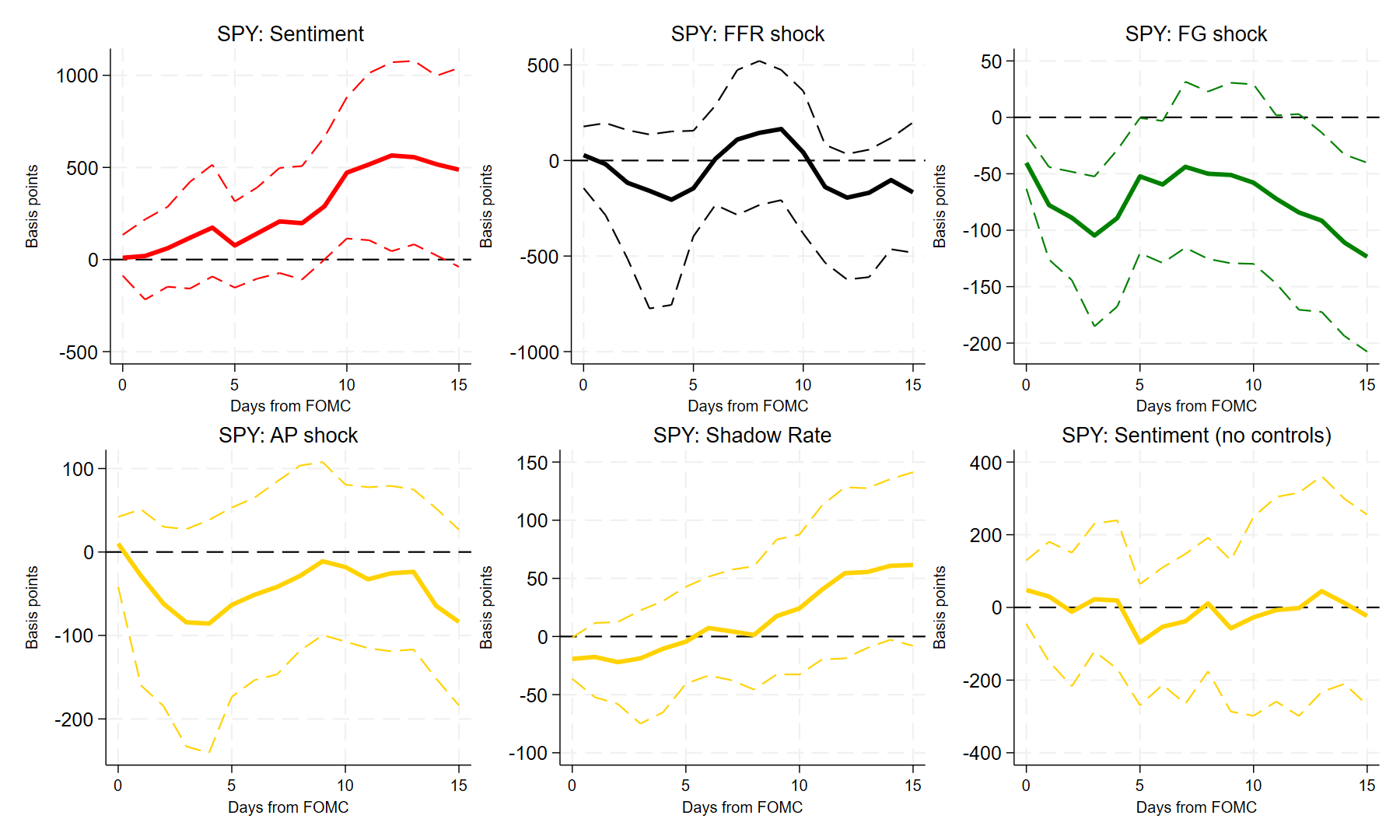}
    \caption{Fine-grained}
    \label{fig:short-a}
  \end{subfigure}%
  \hfill
  \begin{subfigure}[b]{0.5\linewidth}
    \centering
    \includegraphics[width=\linewidth]{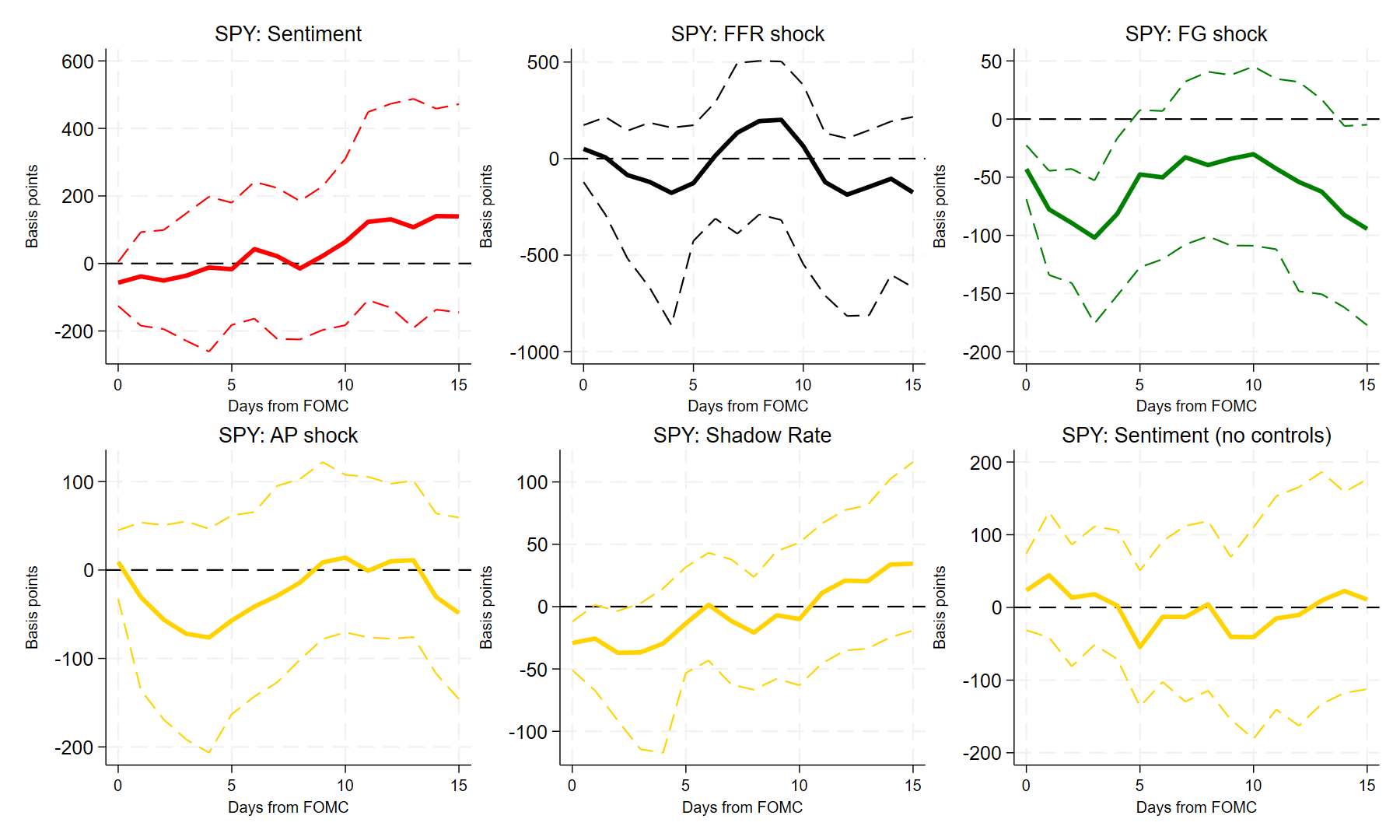}
    \caption{Coarse-grained}
    \label{fig:short-b}
  \end{subfigure}
  \caption{Response of SPY to Sentiment: Different Scale}
  \label{fig:short}
\end{figure*}

\textbf{Press Conference}
The Federal Open Market Committee (FOMC), the Fed's key monetary policymaking body, consists of the Board of Governors' seven members and the presidents of the twelve Reserve Banks. They meet approximately eight times a year to discuss economic conditions and develop strategies for maximum employment and price stability. Since 1994, and regularly after each meeting since May 1999, the FOMC issues statements outlining federal funds rate targets. These statements gained importance after December 2008 when the federal funds rate hit its effective lower bound. To improve transparency, Chairman Ben Bernanke began holding press conferences after four annual meetings starting in April 2011, a practice that expanded to after every meeting since 2019.

Our dataset, sourced from public records, includes conference call videos, audio recordings, and text transcripts from six international banks from 2011 to 2022. We specifically analyze Fed press conference videos from April 27, 2011, to January 26, 2022, covering 47 conferences. Differing from the conversation-level analysis in \cite{thevoice}, we segment our text and audio data into sentences, rather than larger paragraphs, to preserve detailed information. This approach, focusing on sentence-level granularity, provides more precise semantic insights and variation, enhancing our model's understanding and predictive accuracy. Our comparative analysis between sentence-level and conversation-level approaches shows that our fine-grained method is more effective. Besides, it also offers more context than phrases-based or words-based analysis. Our analysis reveals a higher proportion of neutral segments in the data, reflecting the cautious communication style of Fed Chairs, who often use neutral language to avoid unintended policy signals. This tendency towards neutrality is evident in our predictive results, aligning with the Chairs' careful choice of words. 

\subsection{Acoustic and linguistic Analysis}
\textbf{Linguistic Analysis}
Previous studies have demonstrated that fine-tuned pre-trained language models (PLMs) significantly outperform both rule-based models and zero-shot models. Therefore, we employ the RoBERTa-large pre-trained language model as our foundational model, which we further fine-tuned for downstream classification tasks. Numerous studies have successfully fine-tuned PLMs for tasks related to emotion classification. However, our research diverges by aiming to extract policy-related information from text. Drawing inspiration from prior literature \cite{trilliondollar}, we do not employ traditional emotional characteristics such as neutral, positive, and negative. Instead, we fine-tuned the RoBERTa model using a recent open-source dataset \cite{trilliondollar}. This dataset comprises three categories: \textit{Dovish}, \textit{Hawkish}, and \textit{Neutral}. \textit{Dovish} sentences indicate an impending easing of monetary policy, whereas \textit{Hawkish} sentences signal an upcoming tightening of monetary policy. Finally, The model achieved an average testing accuracy of 72.8\% and an average testing F1 score of 72.4\%, using a learning rate of 0.000001.

\textbf{Acoustic Analysis}
In response to the absence of a speech database tailored to policy stances, we adopt the wav2vec2.0 model \cite{wav2vec} which is composed of multi-layer convolutional layers. In order to extract emotion-related acoustic features, the model is fine-tuned to predict the speaker's emotional state on a subset of the Interactive Emotional Dyadic Motion Capture (IEMOCAP) \cite{iemocap} database, a prominent repository for English emotional speech, containing nine emotions. This database includes both scripted and improvisational sessions between two actors, making it a benchmark for English SER systems. It contains 12 hours of audiovisual recordings by ten actors, totaling 10,039 utterances, each annotated by three evaluators. The subset focuses on primary emotional states like neutral, happiness, sadness, and anger. The model achieved an average testing accuracy of 75.3\% across these four emotion categories.


\subsection{Policy Indicator}
After feeding the audio into our model, each segment received a label of predicted sentiment (dovish, neutral or hawkish). We aggregate the labels from sentence level to conference level in a concise way to construct an indicator of the Chair's policy stance during a certain press conference: 
\begin{equation}
    Sentiment = \frac{Dovish-Hawkish}{Dovish+Hawkish}
\end{equation}
The closer this value is to 1, the more dovish in indicates; conversely, the closer this value is to -1, the more hawkish in indicates.
We aggregate the audio segments in a parallel manner:
\begin{equation}
    Voice\,\,Tone = \frac{Positive-Negative}{Positive+Negative}
\end{equation}
The closer this value is to 1, the more positive it indicates; conversely, the closer this value is to -1, the more negative it indicates.
\section{Regression and Impact Analysis}
\subsection{Comparative Analysis of Different Granularity}
\begin{figure}[htbp]
  \centering
  \includegraphics[width=0.48\textwidth]{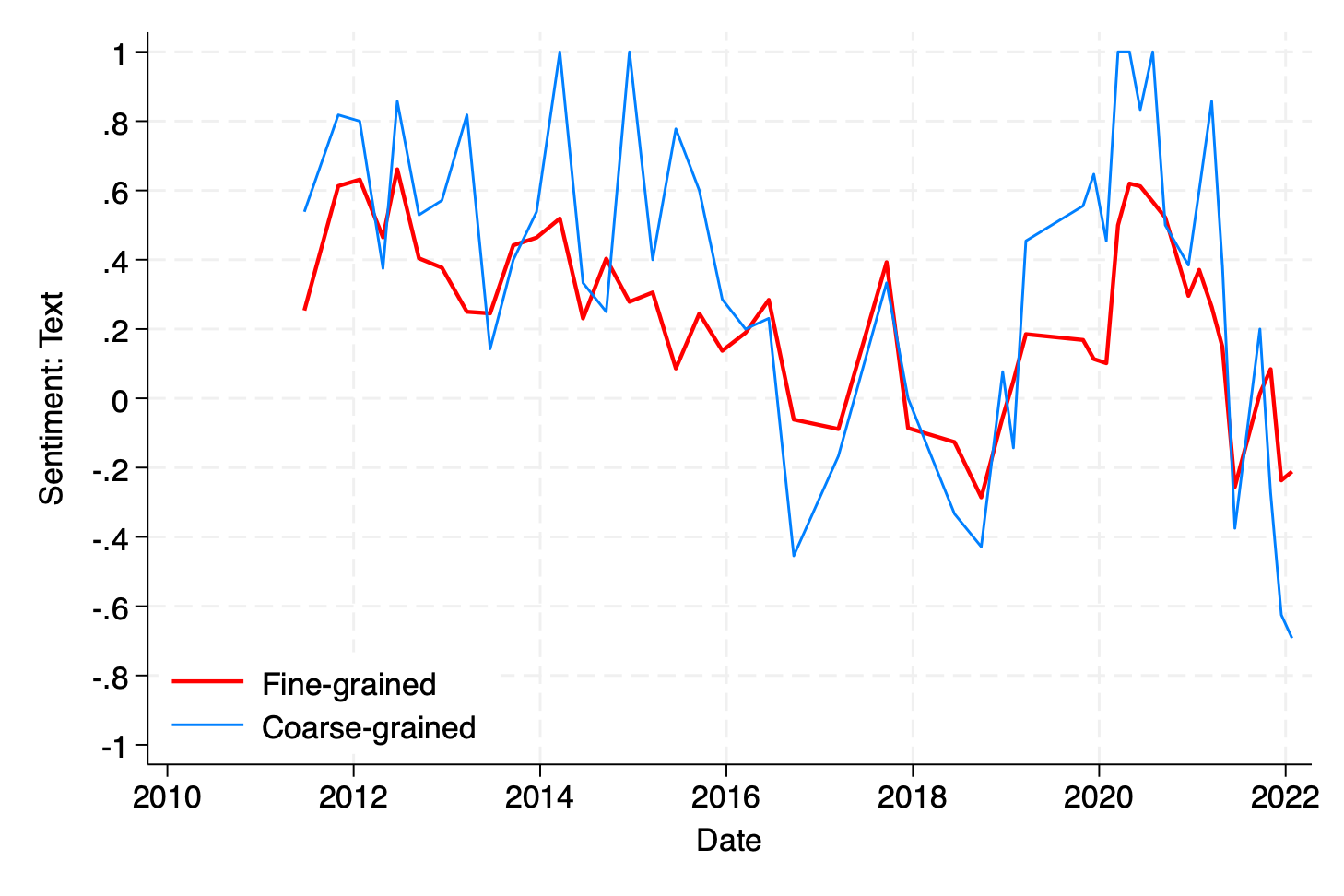}
  \caption{Sentiment Analysis Based on Data of Different Granularities}
  \label{fig:granularity}
\end{figure}

To evaluate the efficacy of the fine-grained method we employ in our framework, We first visualize the distribution of the sentiment indicators' results derived from both fine-grained and coarse-grained methods. The coarse-grained method means we calculate the value of the sentiment indicator based on the conversational level as in \cite{thevoice}. As shown in Fig.~\ref{fig:granularity}, our findings reveal a high correlation in trends of sentiment  across different granularities, suggesting that indicators from varied granularities yield consistent policy implications. Notably, the indicator based on fine-grained data demonstrates smoother trends with fewer extremes compared to its coarse-grained counterpart. This observation aligns with the expectation that as public figures, Chairs tend to maintain prudence and consistency in their communications.
Furthermore, we perform a local-projection regression analysis on U.S. stock prices (SPY) to assess the performance of different granularity levels in predicting financial market volatility. Our results, illustrated in Fig.~\ref{fig:short}, reveal that incorporating a fine-grained sentiment indicator significantly enhances the predictive capability of our framework. The sentiment's coefficient became notably significant at a 10\% confidence level with a 10-day lag. In summary, employing a fine-grained analysis method improves the accuracy and predictability of our framework, contributing to the resolution of critical economic issues.

\subsection{Analysis of Domain-Specific Fine-tuning}
\begin{figure}[ht!]
  \centering 
  \includegraphics[width=0.48\textwidth]{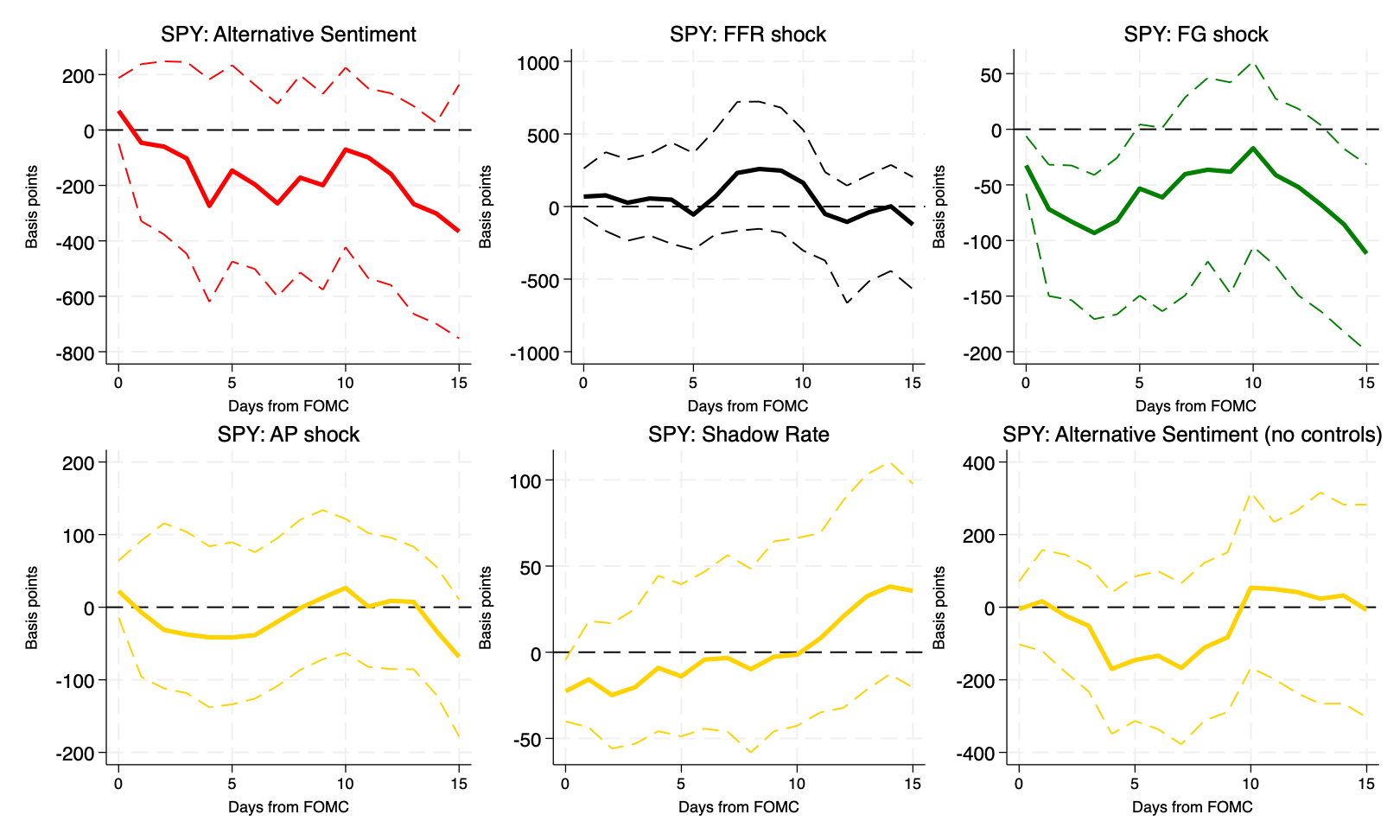}
  \caption{Sentiment Analysis Based on Positive/Negative Labels}
  \label{fig:lbl}
\end{figure}
In this section, we explore the impact of domain-specific fine-tuning. We use RoBERTa \cite{roberta} as a baseline, which has been fine-tuned for three-category emotion recognition (positive, negative, and neutral) \cite{tweeteval}, in contrast to policy stance. We adjust our regression equation to incorporate specific sentiment measures. Our findings show that the financial market's responses vary significantly between Fig.~\ref{fig:short-a} and Fig.~\ref{fig:lbl}. Notably, the sentiment index based on positive and negative classifications proved consistently insignificant, even with control variables, suggesting it is not a reliable predictor of market volatility. However, one unit increase in the sentiment index using dovish-hawkish classifications increases SPY by approximately 500 bp after a 10-day lag, underscoring the value of domain-specific tuning. Consequently, we argue that the lack of a policy-specific speech corpus hindered our ability to achieve the desired analytical results from audio data, as discussed in Sec.~\ref{sec:modality}.

\subsection{Comparative Analysis of Different Communication Scenarios}
\begin{figure}[ht!]
  \centering 
  \includegraphics[width=0.48\textwidth]{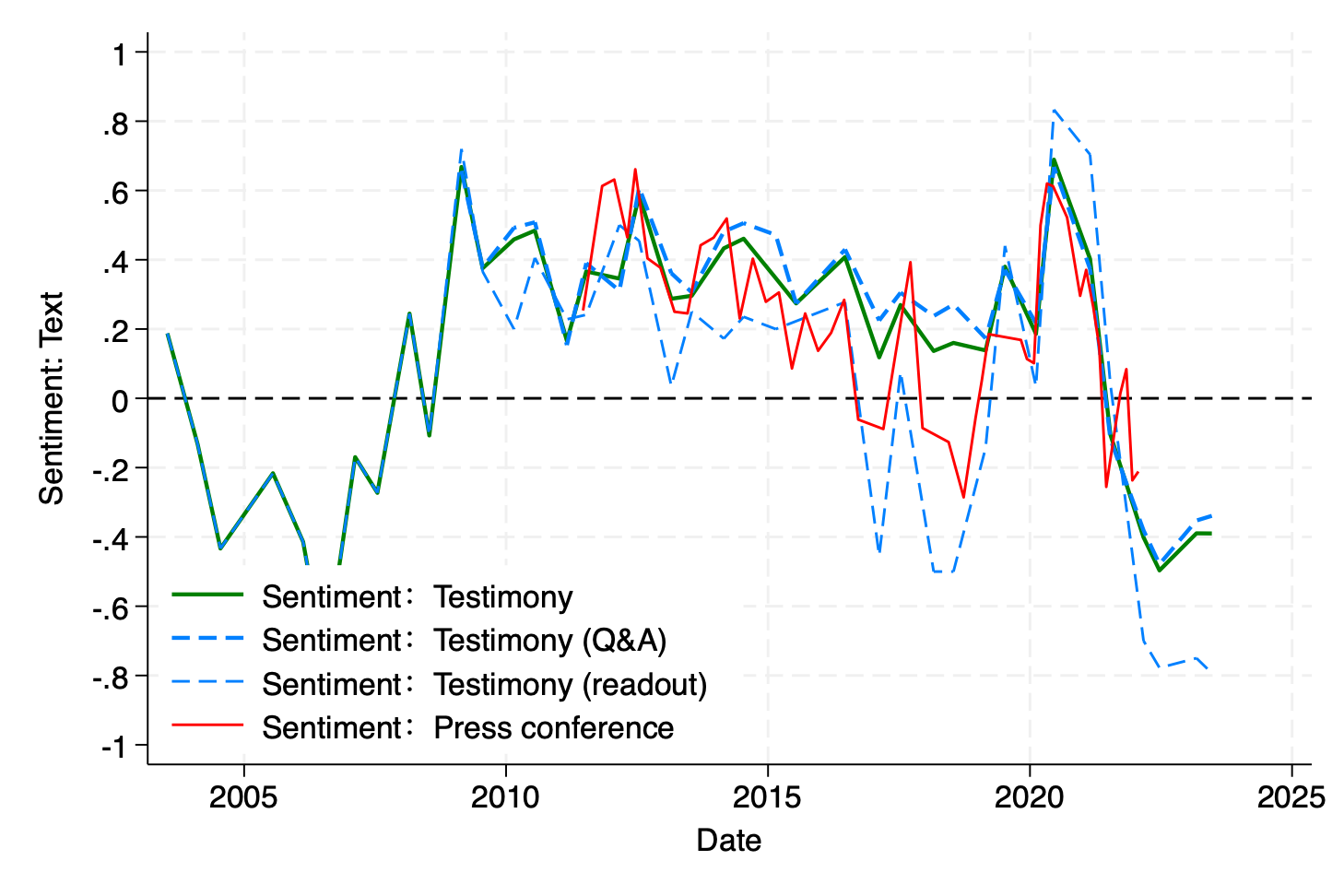}
  \caption{Sentiment Analysis for Press Conference and Monetary Policy Hearings.}
  \label{fig:source-line}
\end{figure}

In this study, we focus on the communication from two scenarios: the Q\&A sessions at the post-FOMC press conferences and the semiannual monetary policy hearings. Both of these two channels play a vital role in the Fed's policy communication: 1) the press conferences serve to prepare the public by shaping their expectations for the upcoming policy decisions (\citet{ehrmann2007timing}; \citet{boguth2019shaping}), 2) the semiannual monetary policy hearings constitute an important communication channel between the Fed and Congress, by providing opportunities for the Fed to convey its policy intentions and economic outlook(\cite{labonte2008monetary}). The dynamics of sentiment orientation are presented in Fig.~\ref{fig:source-line}. We observe that over a given period of time, the inclinations and dynamics of the monetary policy reflected in what the Chair said in both scenarios of meetings remain highly consistent. The analysis results align with the actual economic situation. For example, since 2010, the Fed's monetary policy can be broadly categorized into the following stages: 1) From 2010 to 2014, the Fed supported economic recovery after the Great Financial Crisis with very low interest rates and quantitative easing, which could be reflected with the dovish sentiment result; 2) During 2015 and 2019, the sentiment tended to be a little bit more hawkish as the Fed tapered its asset purchases and gradually increased the federal funds rate; 3) Between 2020 and 2021, in response to the COVID-19 pandemic, the Fed cut rates to near zero and resumed quantitative easing to support the economy, as reflected in Fig.~\ref{fig:source-line} as a jump; 4) After 2021, as inflation pressures mounted, the Fed started discussing tapering asset purchases again, setting the stage for potential rate hikes, thus we can see a giant drop in all the four lines in Fig.~\ref{fig:source-line} accordingly, reflecting that the Chair Powell is getting much more hawkish in every talks he gave. In conclusion, our recognition of the policy stance is basically in line with the literature \citet{bernanke2020new}.

Meanwhile, some systematic differences are also shown in Fig.~\ref{fig:source-line}. The sentiment of policy communication tends to be more neutral at press conferences, as the score of red line is closer to zero. We explain the reason that the Federal Reserve Act requires the Fed chair to report to Congress twice a year on monetary policy, provide explanations, and answer questions from members of Congress. The Chair needs to explain the Fed's policy path in a transparent way so that he can assure Congress that he has been highly responsible, thus the sentiment value for other lines, besides the red one, can be volatile more according to the varying economic fundamentals. On the other hand, when communicating with the public, the Chair would try to be neutral to avoid unnecessary panic in the market. Thus, it's natural that Chairs would sound relatively neutral in the press conferences.

\subsection{Comparative Analysis of Different Modalities}
\label{sec:modality}
\begin{figure}[htbp]
  \centering 
  \includegraphics[width=0.48\textwidth]{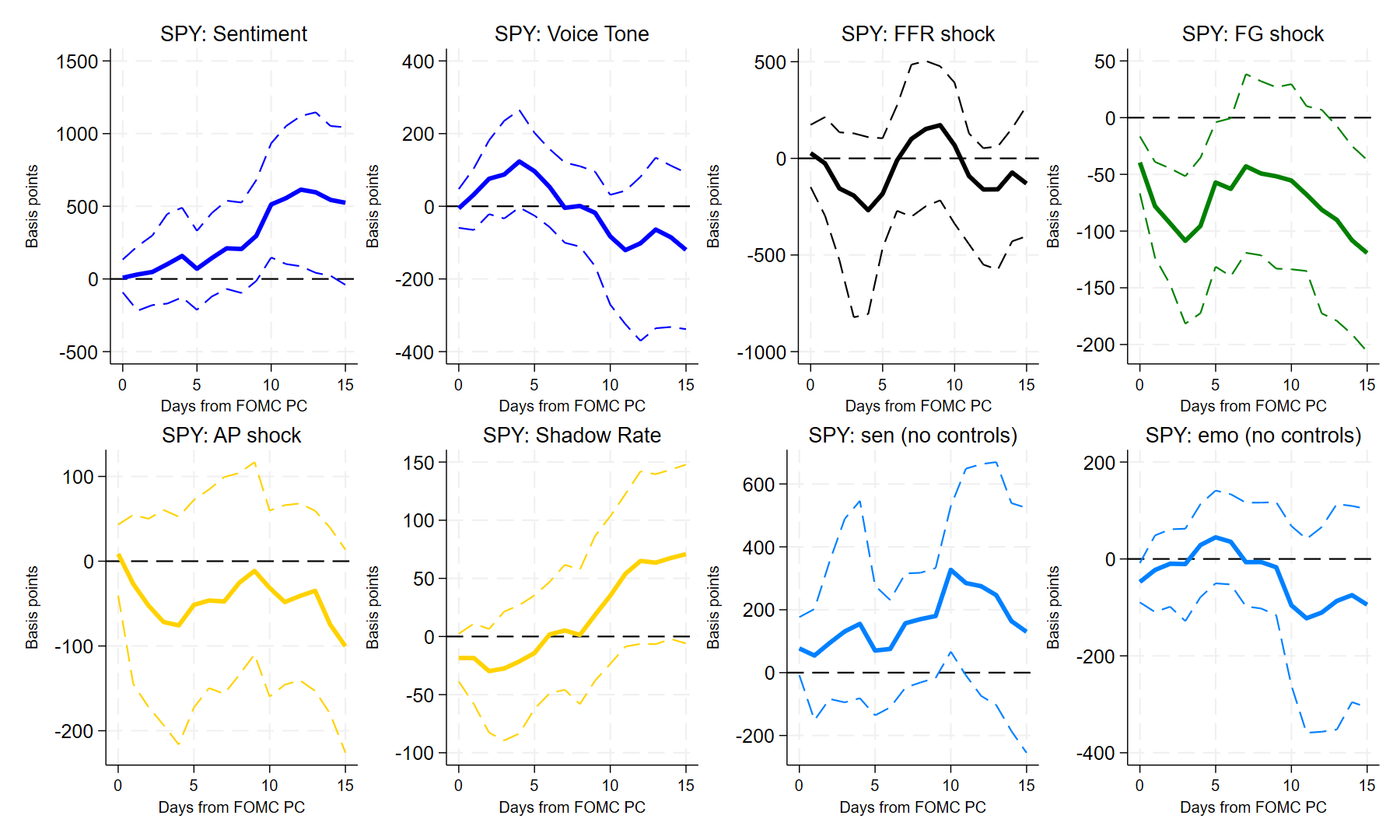}
  \caption[Short title]{Response of SPY to Sentiment and Tone. This figure illustrates how the SPY index reacts to various sentiment and tone changes over time.}
  \label{fig:spy16k}
\end{figure}

To compare the effectiveness of our approach under different modalities, we conduct two parallel empirical regressions on text and audio modalities respectively, as shown in Fig.~\ref{fig:spy16k}. The result reveals that the Voice Tone index, calculated from acoustic sentiment tags, lacked statistical significance in forecasting the financial market. We speculate that the scarcity of speech corpus tailored to the financial sector notably hampers the method's efficacy. Consequently, the model's capacity to discern the underlying sentiment in the Fed Chair's speech is compromised, thereby impeding its predictive power to financial market dynamics. Therefore, it is of significant importance to establish a speech dataset related to the financial sector in the future.

\subsection{Analysis for Cross-border Transmission}
The cross-border transmission of monetary policy is quite an important empirical issue in finance and economics. The existing literature in this field is mainly based on diverse interest-rate indicators to measure monetary policy (\cite{barbosa2018cross}), while the soft information channel represented by the Chair's speeches has been underexplored. This paper considers the impact of the policy stance revealed in the speeches of the Fed Chair on other major economies, including China and the EU. We investigate the variance of three key financial variables, namely the stock price, the monetary-policy-related interest rate and the exchange rate between the national currency and the US dollars.

\begin{figure}[ht!]
  \centering 
  \includegraphics[width=0.48\textwidth]{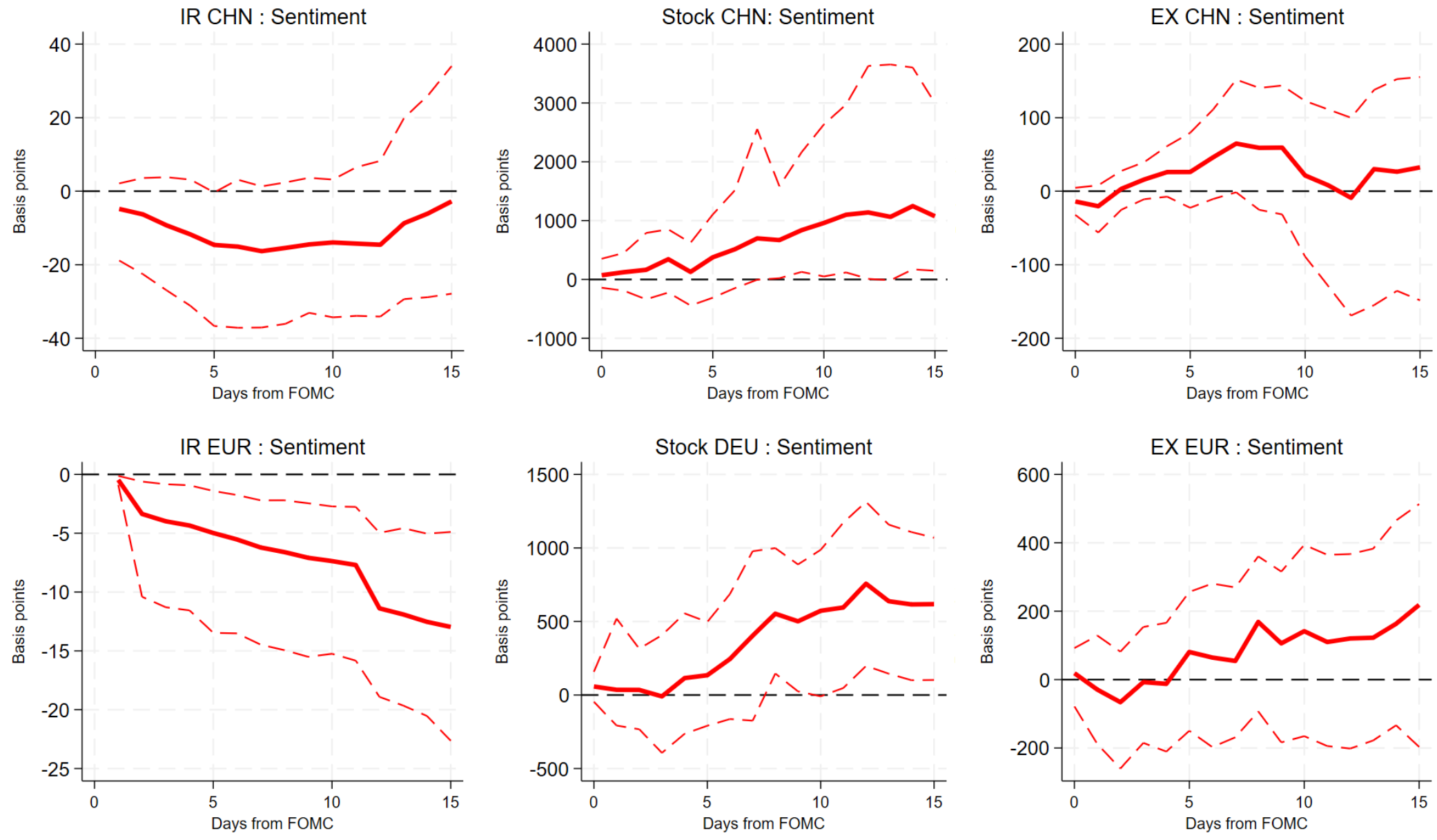}
  \caption[Response of Chinese and European Financial Markets to Sentiment]{Response of Chinese and European Financial Markets to Sentiment: A Comparative Analysis}
  \label{fig:CHN-EUR}
\end{figure}

The impulse response results in Fig.~\ref{fig:CHN-EUR} show that the Fed Chairs' language do have a spillover impact on the financial markets abroad. The sign of the coefficients is the same, while the magnitude diversify across China and the European Union. For China, the sentiment has an immediate impact on China's short-term money market rates, leading to a 15 bp decrease within 10 days; the response of the stock market is slightly delayed, with over 500 bp increase in the stock price in about a week later; however, the sentiment does not show a significant impact on the exchange rate between US dollars and China RMB, which may because after the rapid and flexible adjustment of interest rates, the interest rate parity condition is met again, at which point the exchange rate does not need further adjustment. The responses of financial markets in the EU are basically consistent with China, whether in sign or magnitude. However, the duration of the spillover effect is different. In terms of the responses of the stock prices and interest rates, the effect diminishes in 10 days in China, while it continues to move the corresponding markets in EU in each horizon. In conclusion, the effect of the Chairs' language would spill over to the financial markets in other countries, especially those major economies which are closely linked to the USA in such a globalized world.

\section{Conclusion}
In this paper, we propose a novel, fine-grained monetary policy framework, FMPAF, for analyzing Fed communications from various data sources. We utilize a pre-trained large language model, fine-tuning it with a monetary policy-oriented corpus to transfer knowledge to this specific policy domain. Besides, we employ a pre-trained acoustic model to identify paralinguistic information in the Fed's communications. Our analysis has revealed that domain-specific fine-tuning plays a crucial role in achieving significant results. After applying domain-specific fine-tuning, our method also demonstrates marginal improvements in significance. Furthermore, we verify the spillover effect of the Fed Chairs' language on the international financial market.

\bibliography{aaai24}
\end{document}